\documentclass[10pt,twocolumn,letterpaper]{article}

\usepackage{iccv}
\usepackage{times}
\usepackage{epsfig}
\usepackage{graphicx}
\usepackage{amsmath}
\usepackage{amssymb}
\usepackage{booktabs}
\usepackage{bm}
\usepackage{subcaption} 
\usepackage[accsupp]{axessibility}  

\usepackage[pagebackref=true,breaklinks=true,letterpaper=true,colorlinks,bookmarks=false]{hyperref}

\iccvfinalcopy 

\def\iccvPaperID{11934} 

\ificcvfinal\fi

\begin{document}

\title{Human-Inspired Facial Sketch Synthesis with Dynamic Adaptation}


\author{
	Fei Gao$^1$, 
	Yifan Zhu$^2$, 
	Chang Jiang$^2$, 
	Nannan Wang$^{3*}$ \\
	$^1$Hangzhou Institute of Technology, Xidian University 
	$^2$Hangzhou Dianzi University
	$^3$Xidian University  \\
	{\tt\small fgao@xidian.edu.cn, 
	2961695289@qq.com, 
	jc233@hdu.edu.cn,
	nnwang@xidian.edu.cn }
}


\maketitle
\ificcvfinal\thispagestyle{empty}\fi

{
	\let\thefootnote\relax\footnotetext{* Corresponding Author}
}

\begin{abstract}
Facial sketch synthesis (FSS) aims to generate a vivid sketch portrait from a given facial photo. Existing FSS methods merely rely on 2D representations of facial semantic or appearance. However, professional human artists usually use outlines or shadings to covey 3D geometry. Thus facial 3D geometry (e.g. depth map) is extremely important for FSS. Besides, different artists may use diverse drawing techniques and create multiple styles of sketches; but the style is globally consistent in a sketch. Inspired by such observations, in this paper, we propose a novel \textit{Human-Inspired Dynamic Adaptation} (HIDA) method. Specially, we propose to dynamically modulate neuron activations based on a joint consideration of both facial 3D geometry and 2D appearance, as well as globally consistent style control. Besides, we use deformable convolutions at coarse-scales to align deep features, for generating abstract and distinct outlines. Experiments show that HIDA can generate high-quality sketches in multiple styles, and significantly outperforms previous methods, over a large range of challenging faces. Besides, HIDA allows precise style control of the synthesized sketch, and generalizes well to natural scenes and other artistic styles. Our code and results have been released online at: \url{https://github.com/AiArt-HDU/HIDA}. 
\end{abstract}

\section{Introduction}
\label{sec:intro}

Making computers create arts like human beings, is a longstanding and challenging topic, in the artificial intelligence (AI) area \cite{artai22}. 
To this end, researchers have made great efforts and proposed numerous methods, such as neural style transfer (NST) \cite{Huang2017AdaIN} and image-to-image translation (I2IT) \cite{Isola2017Pix2Pix, Zhu2017CycleGAN}. These methods mainly tackle cluttered image styles, such as oil paintings \cite{Huang2017AdaIN}. In this paper, we are interested in creating artistic sketches from facial photos, which is referred to as face sketch synthesis (FSS) \cite{Wang2013Transductive}. 

For now, there has been significant progress in FSS inspired by the excellent success of Generative Adversarial Networks (GANs) \cite{Isola2017Pix2Pix}. 
Specially, researchers have proposed various techniques, including embedding image prior \cite{Zhang2018IJCAI}, semi-supervised learning \cite{Chen2018Semi}, self-attention/transformer based methods \cite{gao2020incremental, zhu2021sketch,duan2020multi}, hierarchical GANs \cite{Peng2019DeepPGM, Zhang2019TIP, Fan2021FS2K}, composition assistance \cite{gao2020cagan}, and semantic adaptive normalization \cite{li2021genre}, to boost the quality of synthesized sketches. 
However, all these methods merely use 2D appearance or semantic representations of the input photo. 
They may fail to handle serious variations in appearance, such as the pose, lighting, expression, and skin color. 

\begin{figure}
  \includegraphics[width=\linewidth]{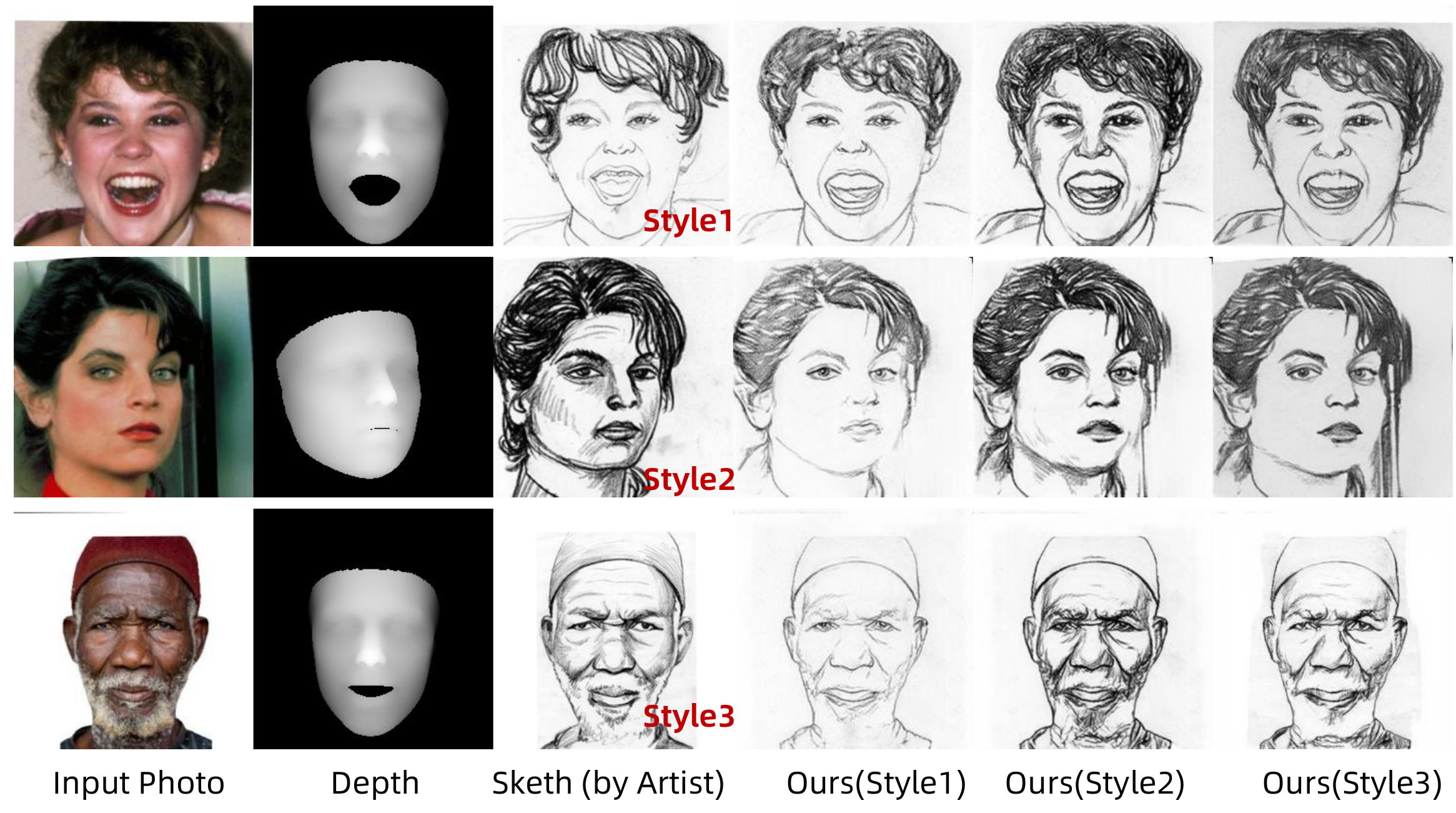}
  \caption{Illustration of facial photos, depth maps, multi-style facial sketches drawn by human artists \cite{Fan2021FS2K}, and the corresponding results synthesized by our method.}
  \label{fig:teaser}
\end{figure}


To tackle this challenge, we propose a novel method, inspired by how human artists draw a sketch. We observe that facial 3D geometry plays a significant role in human artists' drawing process. Besides, a professional human artist considers comprehensive information, including facial 3D geometry, 2D appearance, and the artistic style, to execute a sketch portrait.  
We summarize the drawing methodologies of human artists \cite{li2019im2pencil} into the following four folds:
\begin{itemize}
	\item \textbf{Local 3D geometry conveyor}: First, artists typically use abstract and deformable outlines to characterize major geometry, and use different shading methodologies, e.g. hatching, blending, and stippling, to convey local 3D structures \cite{chan2022informativedraw}. 
	\item \textbf{Local 2D appearance representation}: Second, artists may use different shading or tonal techniques to represent local 2D facial appearance, so as to depict variations in lighting, color, texture, etc.
	\item \textbf{Sketches in diverse styles}: Third, different artists may use diverse drawing methods and create multiple styles of sketches. In other words, they may use divergent textures to represent the same facial area. Fig. \ref{fig:teaser} shows three styles of sketches drawn by artists \cite{Fan2021FS2K}. Obviously, Style1 is extremely abstract and mainly contains sketchy outlines. In contrast, Style3 depicts facial 3D geometry with a lot of shading textures.
	\item \textbf{Globally consistent style}: Finally, the style of pencil-drawing is usually consistent in a single sketch. As shown in Fig. \ref{fig:teaser}, although there are distinct inter-style divergences, the style of pencil-drawing is globally consistent across different regions inside each sketch.
\end{itemize}


Inspired by these observations, we seek to guide the synthesis of sketch portraits by using comprehensive information, including facial 3D geometry and 2D appearance, as well as global style control.  
In the implementation, given a facial photo, we use the depth map to represent its 3D geometry, and use the encoding features to represent its 2D appearance. Afterwards, we combine them with a style map to dynamically modulate deep features for generating a sketch. 
Inspired by the success of SPADE \cite{Park2019GauGAN} in style control \cite{cocosnet} and the local flexibility of dynamic neural networks \cite{han2021dynamic}, we propose to dynamically modulate neuron activations, based on a joint consideration of all these information. 
Such modulation is conducted though both dynamic normalization and activation. 
Specially, we propose a novel dynamic activation function, termed Informative ACON (InfoACON), and a dynamic normalization module, termed DySPADE. 
In addition, we use deformable convolutions \cite{Dai2017Deformable} to align deep features \cite{FaPN2021ICCV} at coarse scales for generating abstract and distinct sketchy outlines. 
Initially, the dynamic adaptation and deformation simulate the flexibility and abstract process of human artists during drawing.


Based on the above mentioned contributions, we build a Human-Inspired Dynamic Adaptation (HIDA) method for FSS.
We conduct experiments on several challenging datasets, including the FS2K \cite{Fan2021FS2K}, the FFHQ \cite{Karras2018StyleGAN}, and a collection of faces in-the-wild. Our method outperforms state-of-the-art (SOTA) methods both qualitatively and quantitatively. Besides, our method allows precise style control and can produce high-quality sketches in multiple styles. Even for faces with serious variations, the synthesized sketches present realistic textures and preserve facial geometric details. In addition, extensive ablation studies demonstrate the effectiveness of the proposed dynamic and adaptive modulation techniques. Finally, our model, although trained for faces, can generate high-quality sketches for natural scenes. 


\section{Related Works}
\label{sec:related}
Our work is related to GANs-based FSS methods. Besides, our method is highly inspired by semantic adaptive normalization \cite{Park2019GauGAN} and dynamic activation \cite{ma2021acon}. 

\textbf{GANs-based FSS.}
\label{ssec:fss}
The latest FSS methods are typically based on GANs \cite{Isola2017Pix2Pix, Gui2020ReviewGAN}, where the mapping from a facial photo to a sketch is modeled as an image-to-image translation task \cite{Isola2017Pix2Pix}.
Some latest methods use 2D semantic information to guide the generation process. For example, Yu et al. \cite{gao2020cagan} propose a stacked composition-aided GANs to boost quality of details. Inspired by the great success of spatially adaptive (de)normalization (SPADE)  \cite{Park2019GauGAN} in semantic image generation, Wang et al. \cite{zhu2021sketch} and Qi et al. \cite{qi2022biphasic} spatially modulate decoding features according to facial parsing masks. Li et al. \cite{li2021genre} propose an enhanced SPADE (eSPADE) by using both facial parsing masks and encoding features for feature modulation. 

Recently, researchers seek to solve the challenge of unconstrained faces by constructing large datasets. Fan et al. \cite{Fan2021FS2K} release a challenging FS2K dataset, which consists of multi-style sketches for faces with diverse variations. 
Nie et al. \cite{nie2021unconstrained} propose a novel WildSketch dataset and a Perception-Adaptive Network (PANet). In PANet, deformable feature alignment (DFA) and patch-level adaptive convolution are used. Different from \cite{nie2021unconstrained}, we analyze the effects of DFA, and only use DFA over the coarse scales. Besides, we propose to dynamically modulate neuron activations based on facial depth and artistic style.



\textbf{Semantic Adaptive Normalization.} 
\label{ssec:spade}
Recently, Park et al. \cite{Park2019GauGAN} propose to modulate deep features based on semantic layouts for semantic image synthesis. In SPADE, deep features are modulated based on semantic layouts. Afterwards, Zhu et al. \cite{zhu2020sean} propose Semantic Region-Adaptive Normalization (SEAN) to control the style of each semantic region individually. 
To boost the efficiency of SPADE, Tan et al. \cite{tan2021diverse} propose a Class-Adaptive (DE)Normalization (CLADE) layer by replacing the modulation networks with class-level modulating parameters. 
All these adaptive normalization layers use 2D semantic maps and show amazing performance in generating photo-realistic images \cite{lv2021learning} and face sketches \cite{gao2020cagan}. 
In this paper, we use pix-wise dynamic activation in the normalization block, so that the modulating parameters would flexibly adapt to local information. 
Experimental results show that the dynamic normalization are essential for detailed synthesis of facial sketches.  

\begin{figure*}[ht]
	\centering
	\includegraphics[width=1\linewidth]{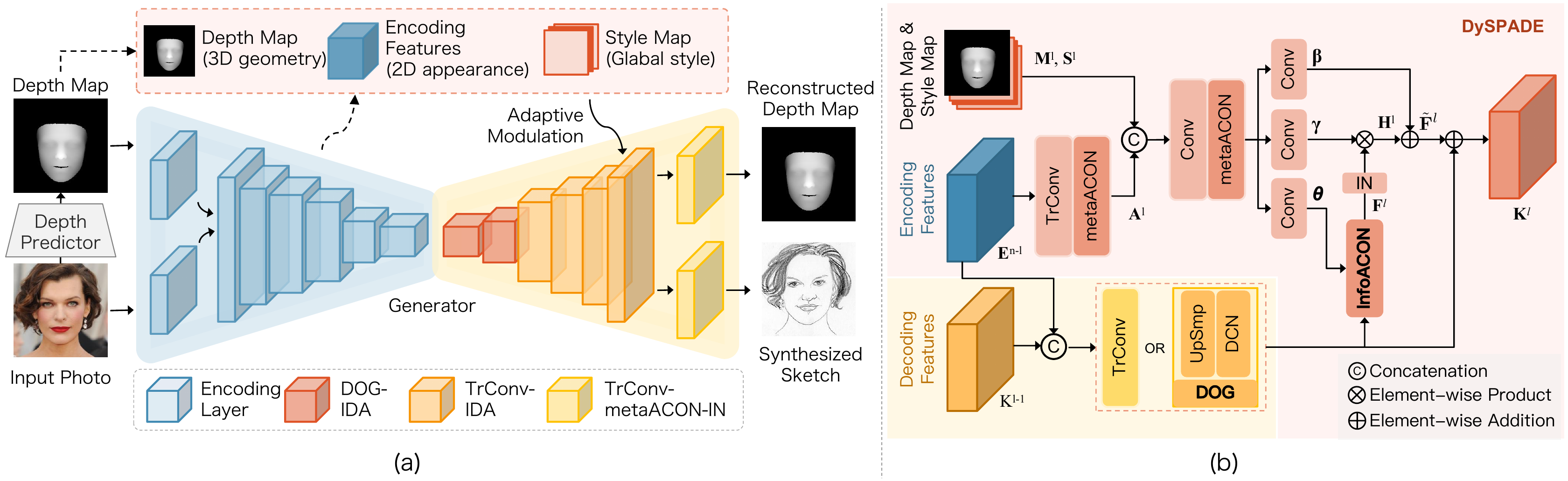} 
	\caption{Pipeline of the proposed \textit{Human-Inspired Dynamic Adaptation} (HIDA) method for facial sketch synthesis. (a) The overall generator architecture, (b) an decoding layer with DySPADE, InfoACON, and DOG.}
	\label{fig:pipeline}
\end{figure*}

\textbf{Dynamic Activations.}
Recently, Chen et al. \cite{dyrelu} propose a Dynamic ReLU (DY-ReLU) function, where parameters in Leaky ReLU are learned from all input elements. Ma et al. \cite{ma2021acon} propose a costumed activation function, termed ACON, which automatically decides whether a neuron is active or not. ACON has several variants, among which the pixel-wise version of metaACON shows remarkable performance. Give a neuron activation $x$, the output of metaACON is formulated as:
\begin{equation}
	y = (p_1 - p_2) \cdot \sigma(\theta(p_1 - p_2)x) + p_2x,
	\label{eq:acon} 
\end{equation}
where $\theta = \sigma(x)$, $\sigma$ is a Sigmoid function, $p_1$, and $p_2$ are learnable parameters.  
In this paper, we use metaACON, instead of ReLU or Leaky ReLU, in part of our networks. Besides, we propose to learn spatially-adaptive parameter $\theta$ according to the 3D geometry, 2D appearance, and global style control. Our activation function proves boosting the performance and allowing precise style control.

\section{The Proposed}
\label{sec:method}


We aim to translate a facial photo $\mathbf{X}$ to a sketch $\mathbf{Y}_s$, in style $s$, drawn by an artist. Here $s=1,2,...,S$ is a style label, $S$ is the total number of styles. 
In this work, we seek to guide the synthesis of sketch portraits by using comprehensive facial information, including both 3D geometry and 2D appearance, as well as the global style control. 
Given a facial photo, we use the corresponding depth map $\mathbf{D}$ to represent its 3D geometry. Afterwards, we combine them with a global style map $\mathbf{S}$ to decode a facial sketch. In this way, our goal is formulated as learning a mapping from $\{\mathbf{X}, \mathbf{D}, \mathbf{S}\}$ to $Y_s$, i.e. $G: \{\mathbf{X}, \mathbf{D}, \mathbf{S}\} \mapsto \mathbf{Y}_s$.

To supervise our model, it is necessary to obtain depth maps for input facial photos. However, it is usually impossible to obtain ground truth depth information in practical applications. 
Therefore we use state-of-the-art (SOTA) depth prediction methods to estimate the depth map of an input facial photo. In practice, we use 3DDFA \cite{3DDFA} as the depth predictor, because it has been widely used and shown excellent performance in various 3D face reconstruction tasks. 

The overall pipeline of our model is as shown in Fig. \ref{fig:pipeline}. It contains an off-line facial depth predictor $P$, a generator $G$, and a patch-wise discriminator $D$. 
In addition to a facial sketch, we enforce $G$ to reconstruct the input depth map $\mathbf{D}$ from features representing the sketch. In this way, the generated sketch $\hat{\mathbf{Y}}_s$ would convey the 3D geometry of $\mathbf{X}$. 
Besides, we boost the capacity of generator by using a dynamic normalization module and a  dynamic activation function. 
Finally, to formulate the abstraction methodology of human artists in drawing sketchy outlines, we propose using deformable convolutions to align features at coarse scales. Details will be introduced bellow.

\subsection{Informative and Dynamic Adaptation (IDA)}
\label{ssec:IDA}

To simulate the drawing methodology of human artist, we first propose a novel \emph{Informative and Dynamic Adaptation (IDA)} module, to modulate deep features based on a combination of the facial depth map $\mathbf{D}$, the style map $\mathbf{S}$, and the appearance representations $\mathbf{A}$, i.e. $\{ \mathbf{D}, \mathbf{S}, \mathbf{A} \}$. Specially, we propose a novel dynamic activation function, termed Informative ACON (InfoACON), and a dynamic normalization module, termed DySPADE. 



\textbf{Informative ACON (InfoACON).}
\label{ssec:dsaa}
The original metaACON function automatically allows whether a neuron is active or not, based on its value, as previously presented in Eq. \ref{eq:acon}. 
During the drawing process, a human artist typically decides whether to draw a stroke or not based on the 3D geometry, 2D appearance, and style type. Inspired by this observation, we propose to learn the parameter $\theta$ in Eq. \ref{eq:acon} from $\{ \mathbf{D}, \mathbf{S}, \mathbf{A} \}$, i.e.
\begin{equation}
		\theta = \sigma ( \phi_\theta(\mathrm{Cat}( \mathbf{D}, \mathbf{S}, \mathbf{A})),
\end{equation}
where $\phi_\theta$ is a two-layer Convolutional network (Fig. \ref{eq:DySPADE}). 
We refer to the modified metaACON function as \textit{Informative ACON} (InfoACON). In our networks, we apply this InfoACON function in all the decoding layers. In this way, the decoder would pixel-wisely decides whether to depict a stroke, or the type of a stroke, in a generated sketch.


\textbf{Dynamic Normalization (DySPADE).}
Following \cite{Park2019GauGAN}, we additionally transform neuron activations by shifting the mean values and scaling the standard deviations, in the instance-wise and channel-wise manner \cite{ulyanov2017IN}.  Different from the original SPADE, we use dynamic activation here to introduce more flexibility on the learned modulating parameters.
Let $\mathbf{F} \in \mathbb{R}^{C \times H \times W}$ denote the input features of the current DySPADE module. $H$, $W$, and $C$ are the height, width and the number of channels. The activation value at site $(c,h,w)$ is modulated as: 
\begin{equation}
	\tilde{f}_{c,h,w} = \gamma_{c,h,w}(\mathbf{D}, \mathbf{S},\mathbf{A}) \frac{f_{c,h,w}-\mu_{c}}{\sigma_{c}} + \beta_{c,h,w}(\mathbf{D}, \mathbf{S},\mathbf{A}),
	\label{eq:DySPADE}
\end{equation}
where $f_{c,h,w}$ and $\tilde{f}_{c,h,w}$ are the input and modulated activation at site $(c, h, w)$, respectively. $\mu_{c}$ and $\sigma_{c}$ are the mean and standard deviation of $f_{c,h,w}$ in the $c$-th channel. 
$\gamma_{c,h,w}(\mathbf{D}, \mathbf{S},\mathbf{A})$ and $\beta_{c,h,w}(\mathbf{D}, \mathbf{S},\mathbf{A})$ are learned scale and bias parameters at site $(c,h,w)$. 



As shown in Fig. \ref{fig:pipeline}, we use a two-layer and three-branched Convolutional network to predict the parameter $\theta$ in InfoACON, and the modulating parameters $\bm{\gamma}$ and $\bm{\beta}$ in DySPADE. To improve the flexibility of the adaptation block, we use metaACON (Eq. \ref{eq:acon}) \cite{ma2021acon} instead of ReLU, after the first Convolutional layer. In this way, the modulating factors would pixel-wisely adapt to an integration of the facial 3D geometry, 2D appearance, and global style.  

In IDA, the activation at each position is modulated according to a joint consideration of local facial 3D geometry, appearance, and artistic style. This mechanism is consistent with the drawing methodology of human artists. To execute a facial sketch, an artist usually uses diverse textures to represent 3D geometry or illustration variations. Besides, the style of all pencil strokes are consistent inside a single sketch. As a result, IDA is promising to produce realistic sketchy textures in globally consistent style.

\subsection{Deformable Outline Generation (DOG)}
\label{ssec:dfa}

Human artists usually draw abstract lines to capture facial geometric structures, such as the boundaries of facial organs, and facial mood. To this end, the resulting outlines typically convey such structures abstractly, instead of pixel-wisely tracing them. In other words, there are geometric deformations between the input photo and the sketches drawn by artists. 
To simulate such an abstraction drawing methodology, we propose to align decoding features at coarse scales. In this way, the generated sketches would present abstract and distinct outlines, instead of scattered outlines with a lot of subtle variations.
 
In practice, we use deformable convolution (DCN) \cite{Dai2017Deformable} instead of standard Transposed Convolution over the first and second decoding layers. 
As will be presented in the ablation study (Section \ref{ssec:exp_ablation}), this deformable outline generation (DOG) module significantly boosts the clarity of generated outlines. Besides, DOG enables the network produce abstract sketches (e.g. Style1 in the FS2K dataset), which contains a sparse set of sketchy line drawings.

\subsection{Overall Generator Architecture}
\label{ssec:gen}

Our generator follows the U-Net architecture \cite{Isola2017Pix2Pix} in whole. 
In the encoder, the facial photo $\mathbf{X}$ and the depth map $\mathbf{D}$ are first fed into a Convolutional layer, separately. Afterwards, the corresponding feature maps are concatenated and fed into the following encoding layers. Each encoding layer follows a Conv-metaReLU-IN architecture, and down-samples the size of feature maps by 1/2. The encoding features are adopted as appearance representations, $\mathbf{A}$.


In the decoder, we expand an DySPADE block to every decoding layer, except the last one. 
Fig. \ref{fig:pipeline} illustrates the pipeline of a decoding layer with DySPADE. 
Over the $l$-th decoding layer, let $\mathbf{D}^{l}$ be the corresponding depth map, $\mathbf{S}^{l}$ the style map, and $\mathbf{A}^{l}$ the appearance features.
We down-sample the original depth map $\mathbf{D}$ to $\mathbf{D}^{l}$ by building a Gaussian Pyramid, and expand the one-hot style vector $\mathbf{s}$ to $\mathbf{S}^{l}$. 
Besides, we obtain $\mathbf{A}^l$ by upsampling $\mathbf{E}^{l-1}$ through a Transposed-Convlotional (TrConv) layer, followed by a metaACON activation layer.
We finally apply the residual connection to obtain the output of the $l$-th decoding layer: 
\begin{equation}
	\mathbf{K}^{l} = \mathbf{H}^l \oplus \tilde{\mathbf{F}}^l, \text{~with~} \tilde{\mathbf{F}}^l = \mathrm{DySPADE}(\mathbf{F}^l), 
\end{equation}
where $\oplus$ denotes element-wise addition. 
$\mathbf{H}^l$ is the initial upsampled feature map, output by a DOG layer (over the $1^{st}$ and $2^{nd}$ decoding layer) or a TrConv layer (over the rest layers).
$\mathbf{K}^{l}$ is fed into subsequent layers for generating final predictions.

\subsection{Loss Functions}
\label{ssec:loss}
To train our model, we use the following loss functions. 

\textbf{Geometric loss.}
First, we use a geometric constraint to supervise depth reconstructions from features of sketches.
The geometric loss is the L2 distance between the input depth map and the reconstructed one: 
\begin{equation}
	\mathcal{L}_{geo} = \Vert \hat{\mathbf{D}} - \mathbf{D} \Vert_2^2.
\end{equation}

\textbf{Textural loss.}
The synthesized sketch $\hat{\mathbf{Y}}_s$ should present similar textures as that drawn by an artist $\mathbf{Y}_s$. In this work, we constrain $\hat{\mathbf{Y}}_s$ and $\mathbf{Y}_s$ to have similar pixel-wise adjacent correlations \cite{li2021genre}. To this end, we calculate their gradients by using the Sobel operator, and calculate the average Cosine distance between them. Let $\mathbf{g}_{i,j} = [g^x_{i,j}, g^y_{i,j}]^T$ denote the $x$-directional and $y$-directional gradients of $\mathbf{Y}_s$ at site $(i,j)$; and $\mathbf{f}_{i,j} = [f^x_{i,j}, f^y_{i,j}]^T$ the corresponding gradients in $\hat{\mathbf{Y}}_s$. 
The textural loss is formulated as: 
\begin{equation}
	\mathcal{L}_{tex} = \frac{1}{MN} \sum_{i,j}  \frac{\mathbf{g}_{i,j}^T \mathbf{f}_{i,j}}{\Vert \mathbf{g}_{i,j} \Vert \cdot \Vert \mathbf{f}_{i,j} \Vert},
\end{equation}
where $\Vert \cdot \Vert$ denotes the magnitude of a vector, $M$ and $N$ are the width and height of the sketch.

\textbf{Pixel loss.}
In addition, we use the pixel-wise reconstruction loss between the synthesized sketch $\hat{\mathbf{Y}}_s$ and the target sketch $\mathbf{Y}_s$, i.e. 
\begin{equation}
	\mathcal{L}_{pix} = \Vert \hat{\mathbf{Y}_s} - \mathbf{Y}_s \Vert_1.
\end{equation}

\textbf{Adversarial loss.}
Finally, we use adversarial loss to measure whether a pair of depth map and synthesized sketch is real or fake. Here, we use the Cross Entropy loss, i.e. 
\begin{equation}
	\mathcal{L}_{adv} = -\log D(\mathbf{D}, \mathbf{Y}_s) 
	- \log (1-D(\hat{\mathbf{D}}, \hat{\mathbf{Y}}_s)).
\end{equation}

\textbf{Full objective.}
We use a combination of all the aforementioned losses as our full objective:
\begin{equation}
	\mathcal{L}_{all} = \mathcal{L}_{adv} + \lambda_1 \mathcal{L}_{pix} + \lambda_2 \mathcal{L}_{tex} + \lambda_3 \mathcal{L}_{geo},
\end{equation}
where $\lambda_1$, $\lambda_2$, and $\lambda_3$ are weighting factors. We train the generator $G$ and the discriminator $D$ in an alternative manner, to minimize $\mathcal{L}_{all}$. 


\section{Experiments}
\label{sec:exp}
We present a thorough experimental comparison on the challenging FS2K dataset \cite{Fan2021FS2K}. Besides, we conduct a series of ablation study to analyse impacts of the proposed DySPADE, InfoACON, and DOG modules.

\begin{figure*}
	\centering
	\includegraphics[width=1\linewidth]{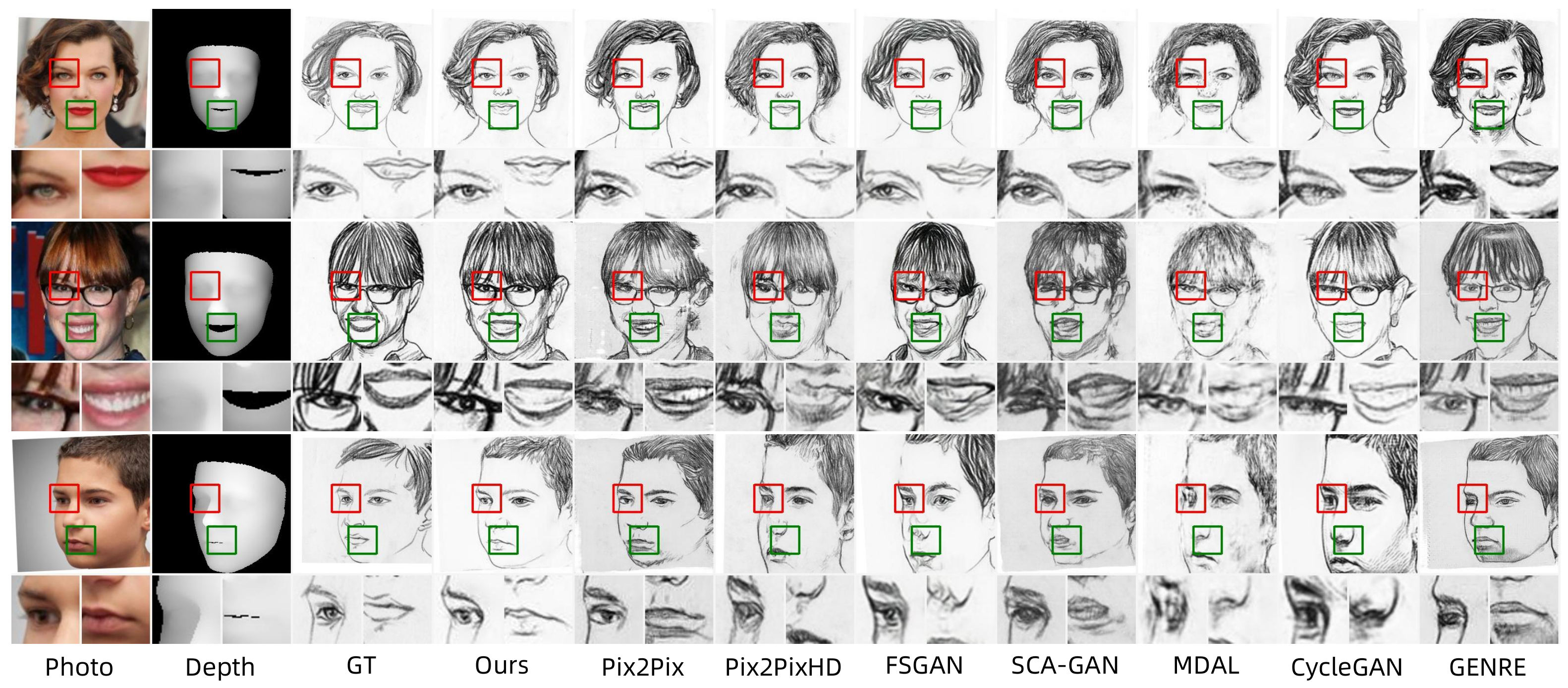}
	\caption{Comparison with SOTAs on the FS2K dataset.}
	\label{fig:fs2k}
\end{figure*}

\subsection{Experimental Settings}
\label{ssec:exp_setting}

\textbf{Data.}
\label{ssec:exp_data}
We conduct experiments on the challenging FS2K dataset. The FS2K dataset is the largest publicly released FSS dataset, consisting of 2,104 photo-sketch pairs from a wide range of image backgrounds, skin colors, sketch styles, and lighting conditions. These sketches are mainly in three styles. Following standard settings \cite{Fan2021FS2K}, we have 1,058 photo-sketch pairs for training, and 1,046 pairs for testing. For each style, we have 357/351/350 training pairs, and 619/381/46 testing pairs, from Style1 to Style3, respectively. All the images are aligned and resized to $250 \times 250$. In the inference stage, we use the same style of sketch as the ground truth in default.  
In addition, we collect a number of challenging Faces in-the-wild from FFHQ \cite{Karras2018StyleGAN} and Web. We align and resize these images in the same way as those in the FS2K dataset. 



\textbf{Comparison Methods}
\label{ssec:exp_compmethod}
In this section, we compare our method with various state-of-the-art (SOTA) ones, including FSGAN \cite{Fan2021FS2K}, GENRE \cite{li2021genre}, SCA-GAN \cite{gao2020cagan}, and MDAL \cite{Zhang2019MAL}. Besides, we compare with several advanced GANs, including CycleGAN \cite{Zhu2017CycleGAN}, Pix2Pix \cite{Isola2017Pix2Pix}, and Pix2PixHD \cite{Wang2017Pix2PixHD}. We use results and codes of these methods released by the corresponding authors \cite{Fan2021FS2K}. All these methods and ours follow the same experimental settings.

\textbf{Criteria}
\label{ssec:exp_criteria}
In this work, we choose four performance indices as the criteria, i.e. the \textit{Fr\'{e}chet Inception distance} (FID) \cite{Heusel2017FID}, \textit{Learned Perceptual Image Patch Similarity} (LPIPS) metric \cite{zhang2018lpips}, \textit{Structure Co-Occurrence Texture} (SCOOT) metric \cite{fan2019scoot}, and \textit{Feature Similarity Measure} (FSIM) \cite{Zhang2011FSIM}.  
Lower values of FID and LPIPS indicate higher realism of synthesized sketches. In contrast, greater values of SCOOT and FSIM generally indicate higher similarity between a synthesized sketch and the corresponding sketch drawn by an artist. We here report the average LPIPS, SCOOT, and FSIM values across all the test samples, respectively. In the following sections, $\downarrow$ indicates that lower value is better, while $\uparrow$ higher is better.

\textbf{Implementation Details}
\label{ssec:implement}
We implemented our model in PyTorch. All experiments are performed on a computer with a Titan 3090 GPU. We use a batch size of 4, a learning rate of $1e-4$. We use the Adam Optimizer, and train the model for 800 epochs on the training set. Our code will be released after peer review. 


\subsection{Qualitative Comparison with SOTAs}
\label{ssec:exp_quali}

We further qualitatively compare with SOTA FSS methods. Fig. \ref{fig:fs2k} illustrates synthesized sketches on the FS2K dataset. Although our method can generate multiple styles of sketches, here we only show the synthesized sketch in the same style as the ground truth. 
For the face in constrained condition (the first row), most methods successfully generate a quality sketch. For the face with extreme lighting condition (the second row) or pose variation (the third row), most synthesized sketches present unpleasant geometric deformations and fail to precisely reproduce the style. 
Although sketches generated by CycleGAN seems acceptable, the textures aren't like pencil-drawings. 
The sketches generated by FSGAN show the same styles as the ground truths, since FSGAN contains a style control module. However, these sketches show unpleasant structural distortions. This might be caused by the geometric deformations between facial photos and free-hand sketches drawn by artists, in the training data. 
GENRE successfully produces quality sketches, but they are all almost in the same style, since no style information is considered in GENRE. 

In contrast, our HIDA generates high-quality sketches in all three styles. Specially, our synthesized sketches preserve the geometries of input faces. This implies that, HIDA doesn't overfit to the training samples and combats geometric deformations. We achieve such success mainly due to the informatively adaptive normalization module, i.e. DySPADE, and the constraint of reconstructing the input depth. 
Besides, our synthesized sketches present the same style of strokes as the corresponding ground truths. The drawing textures are consistent inside each sketch. The style consistency demonstrates the effectiveness of our global style control mechanism through DySPADE and InfoACON. Based on all these observations, we conclude that our HIDA model can generate high-quality and style-consistent sketches.


\begin{table}
	\centering
	\caption{Comparison with SOTAs on the FS2K dataset.}
	\label{tab:pfm_fs2k}
	\renewcommand\arraystretch{1.2}
	\resizebox{0.45\textwidth}{!}{
		\begin{tabular}{l|cccc}
			\toprule
			&	FID$\downarrow$	&	LPIPS$\downarrow$	&	SCOOT$\uparrow$	&	FSIM$\uparrow$	\\
			\midrule	
			Pix2Pix \cite{Isola2017Pix2Pix}	&	\underline{18.34}	&	0.304	&	0.493	&	0.541 	\\
			Pix2PixHD \cite{Wang2017Pix2PixHD}	&	32.03	&	0.468	&	0.374	&	0.531 	\\
			CycleGAN \cite{Zhu2017CycleGAN}	&	26.49	&	0.505	&	0.348	&	0.501 	\\
			MDAL \cite{Zhang2019MAL}	&	50.18	&	0.492	&	0.355	&	0.530 	\\
			SCA-GAN \cite{gao2020cagan}	&	39.63	&	0.305	&	\textbf{0.600} 	&	\textbf{0.782} 	\\
			FSGAN \cite{Fan2021FS2K}	&	34.88	&	0.483	&	0.405	&	\underline{0.610} 	\\
			GENRE \cite{Park2019GauGAN}	&	20.67	&	\underline{0.302}	&	0.483	&	0.534 	\\
			HIDA (Ours)	&	\textbf{15.06}	&	\textbf{0.263}	&	\underline{0.575}	&	0.551 	\\
			\bottomrule
		\end{tabular}
	}
\end{table}

\begin{figure}
	\centering
	\includegraphics[width=1\linewidth]{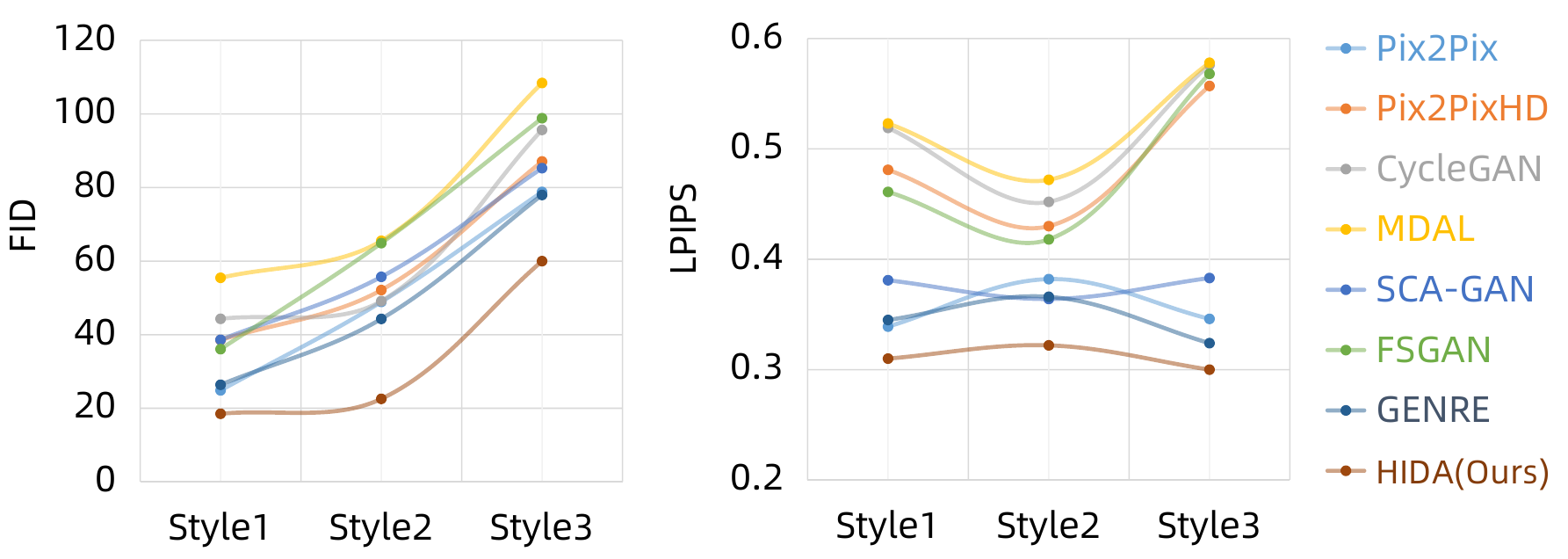}
	\caption{FID and LPIPS values w.r.t. each style in FS2K.}
	\label{fig:fs2kstyle}
\end{figure}

\subsection{Quantitative Comparison with SOTAs}
\label{ssec:exp_quanta}

\textbf{Overall Performance.}
Table \ref{tab:pfm_fs2k} shows the quantitative performance criteria of each method on the whole FS2K testing dataset. 
Obviously, our method achieves the lowest FID and LPIPS values. 
In contrast to previous benchmark method, FSGAN, our HIDA dramatically decrease both FID and LPIPS by about 20 and 0.22, respectively. 
Besides, compared to SOTA 2D-semantic driven methods, i.e. SCA-GAN and GENRE, HIDA decreases FID by about 24 and 5, respectively. HIDA also decreases LPIPS by about 0.04, i.e. 10\% relatively. 
Such dramatic decreases of both FID and LPIPS mean that our method produces the most realistic sketches in terms of style and stroke.

In addition, HIDA achieves the second best value of SCOOT, which is significantly better than FSGAN and GENRE, but slightly lower than SCA-GAN. 
Such a high value of SCOOT means that the sketches produced by our method are similar to those drawn by artists in terms of structure and textures. 
Finally, HIDA achieves the third best FSIM value. Recall that there are geometric deformations between facial photos and sketches drawn by artists. Thus an excessively high value of FSIM might indicate the potential that: a FSS model overfits to the training data, and cannot precisely preserve facial structures in the translation process. Correspondingly, as shown in Fig. \ref{fig:fs2k}, both SCA-GAN and FSGAN produce deformable sketches. In contrast, HIDA preserves the structure of input faces. 

\textbf{Performance on Each Style.}
We further analyse the performance of FSS methods on each style subset. Since both FID and LPIPS measure the realism of synthesized sketches in terms of style and textures, we report them in Fig. \ref{fig:fs2kstyle}. Obviously, our HIDA model consistently achieves the lowest FID and LPIPS values, across all the styles. Especially, our method significantly outperforms previous SOTA method, FSGAN, according to both criteria. Such distinct superiority over existing methods demonstrates that our method effectively learns the style information and allows precise control over the style of synthesized sketches. 

\subsection{User Study}
\label{ssec:exp_subj}

We further conduct a series of subjective study to evaluate the performance of HIDA, in contrast to existing methods. Specially, we have 10 participators, all of whom are not professional artists. For each participator, we show them 1,000 randomly selected samples from the testing set in FS2K. Each time, we show a facial photo, the corresponding sketch drawn by an artist, and 8 synthesized sketches produced by different methods. Participators are requested to choose the best sketch, according to (1) the similarity between a synthesized sketch and the ground truth, and (2) the quality of a sketch, based on their own preferences. Finally, we collect totally 10,000 preference labels. 

\begin{figure}
\centering
\includegraphics[width=1\linewidth]{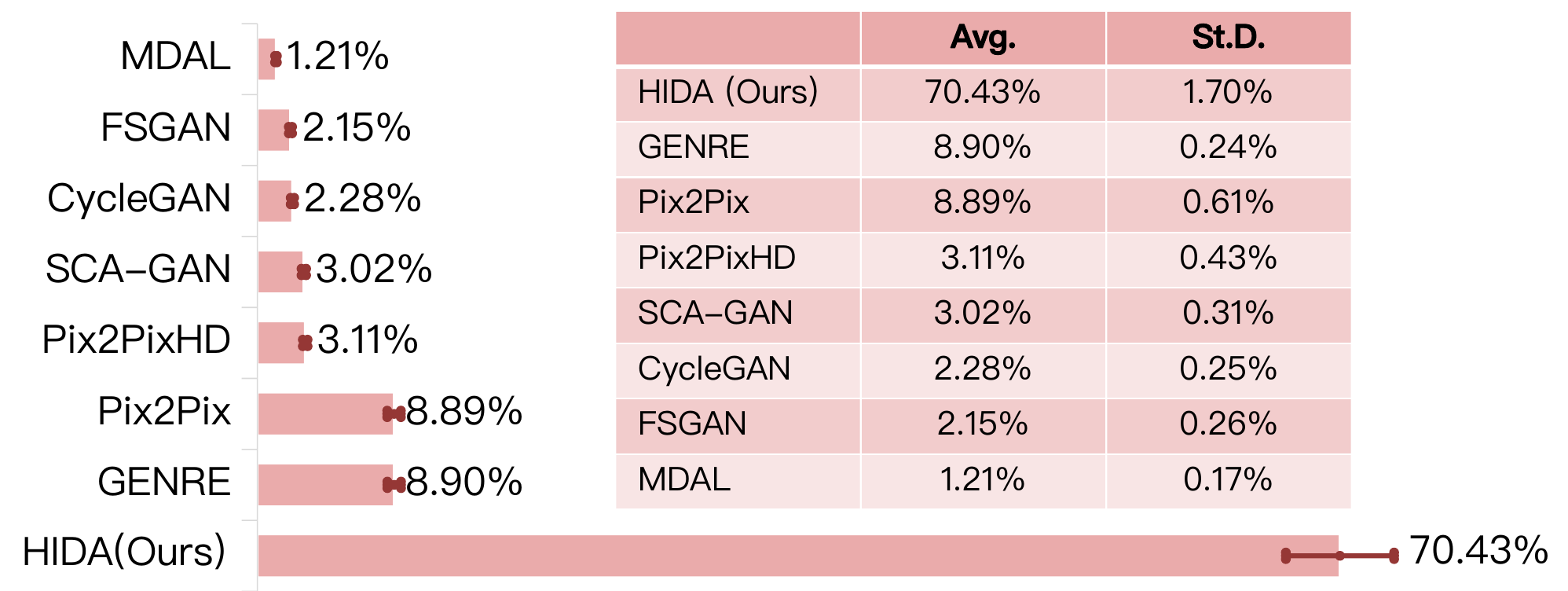}
\caption{Average subjective preference percent  (\textit{Avg.}) and the standard deviations among different subjects (\textit{St.D.}).}
\label{fig:subject}
\end{figure}

Fig. \ref{fig:subject} shows the average preference percent about each model, and the standard deviation among different participants. Obviously, our method dramatically outperforms all the other methods. In average, subjective participators think our model generates the best sketch over 70\% of facial photos. 
The subjective comparison result demonstrate that our method significantly outperforms SOTAs in generating high-quality and style-specific facial sketches. In addition, the sketches synthesized by our HIDA model meet the preference of most users.

\subsection{Ablation Study}
\label{ssec:exp_ablation}

We first conduct a series of ablation study on the FS2K dataset. To this end, we build several model variants, by gradually adding different modules to the base model, i.e. Pix2Pix \cite{Isola2017Pix2Pix}. The modules we aims to analyse include the use of depth map $\mathbf{D}$ as auxiliary input, the DySPADE transformation, the InfoACON function, and the DOG layer.

\begin{figure}
	\centering
	\includegraphics[width=\linewidth]{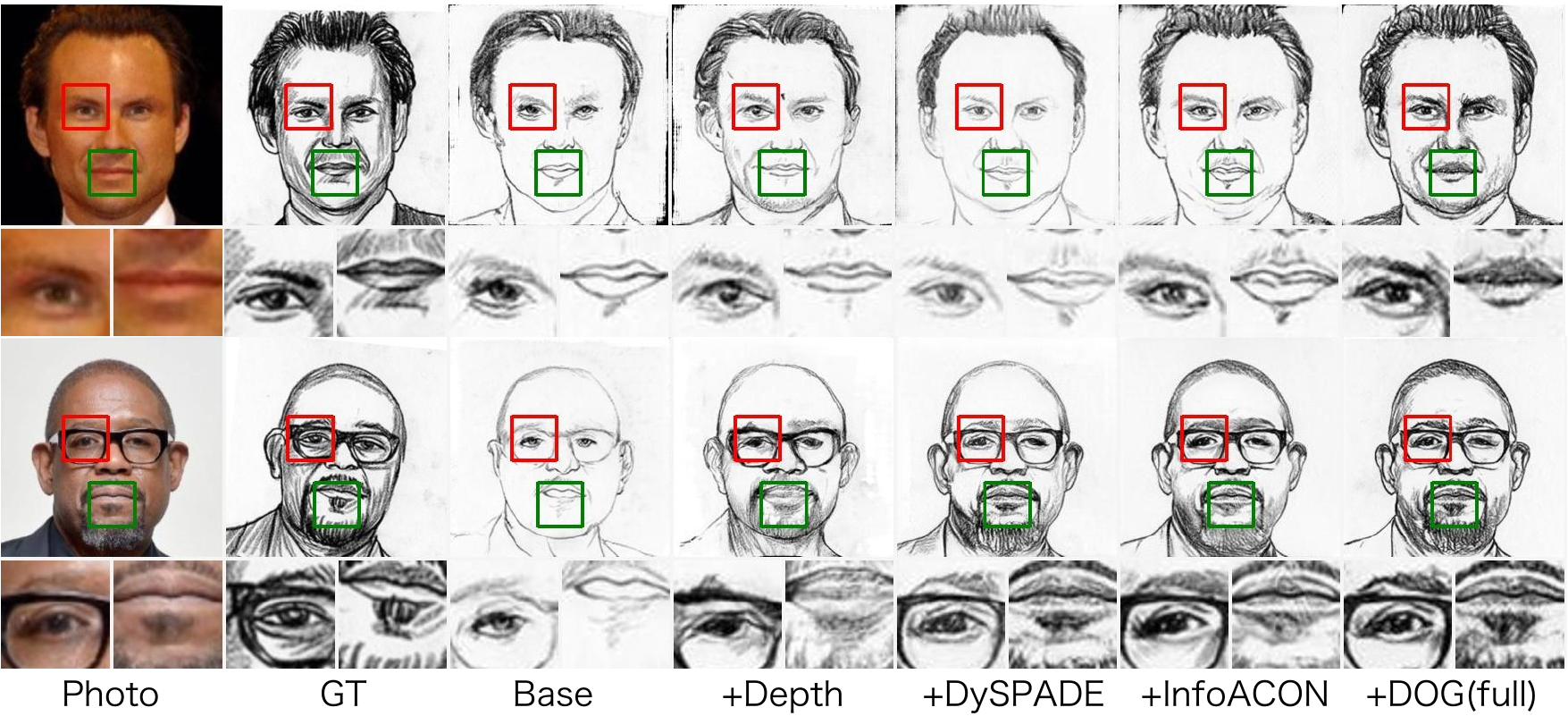}
	\caption{Comparison between different model variants in the ablation study, on the FS2K dataset.}
	\label{fig:exp_ablation}
\end{figure}

\textbf{Qualitative Analysis.}
Fig. \ref{fig:exp_ablation} illustrates sketches produced by these model variants. 
The second column shows the depth maps predicted by 3DDFA. These maps convey well with the corresponding facial geometry in general. 
The third column shows sketches generated by the base model (i.e. Model-A). Obviously, these sketches occasionally show chaotic facial structures. Besides, there is no distinct difference between the generated two sketches in terms of style.  
In contrast, using the DySPADE module (i.e. Model-C) enables the model precisely preserving tiny facial structures. For example, the shapes of eyebrows in both examples become consistent between the synthesized sketches and the input photos. 

If we further use the InfoACON function in the decoder (i.e. Model-D), the generator produces more details. For example, the textures precisely present the 3D structure of lips. Besides, the major boundary of eyeglass is generated. The synthesized sketches of these two examples also show different types of strokes over the same semantic regions, e.g. lips. 
Finally, using DOG (i.e. the full model) enables the model generating abstract and distinct outlines. For example, the result in the top row is consistent with Style1 in terms of line drawings. All the other model variants produce obvious rendering textures to present 3D geometry. 
Such comparisons demonstrate our motivation of using DOG to simulate the abstraction process of human artists.

\begin{table}
	\centering
	\tabcolsep=1.5pt
	\caption{Quantitative results of the ablation study on the FS2K dataset.}
	\label{tab:ablation}
		\begin{tabular}{l|cccc}
			\toprule
			&	FID$\downarrow$	&	LPIPS$\downarrow$	&	SCOOT$\uparrow$	&	FSIM$\uparrow$	\\
			\midrule
			Base ~~(Model-A) 	&	18.34	&	0.304	&	0.493	&	0.541	\\
			+ Depth ~(Model-B)	&	20.1	&	0.307	&	0.489	&	\underline{0.543} \\
			+ DySPADE ~(Model-C)	&	18.30	&	0.298	&	0.479	&	0.539	\\
			+ InfoACON ~(Model-D)	&	\underline{17.41}	&	\underline{0.291}	&	\underline{0.493}	&	0.536	\\
			+ DOG (full)	&	\textbf{15.06}	&	\textbf{0.263}	&	\textbf{0.575}	&	\textbf{0.551}	\\
			\bottomrule
		\end{tabular}
\end{table}

\begin{figure*}
	\centering
	\includegraphics[width=0.9\linewidth]{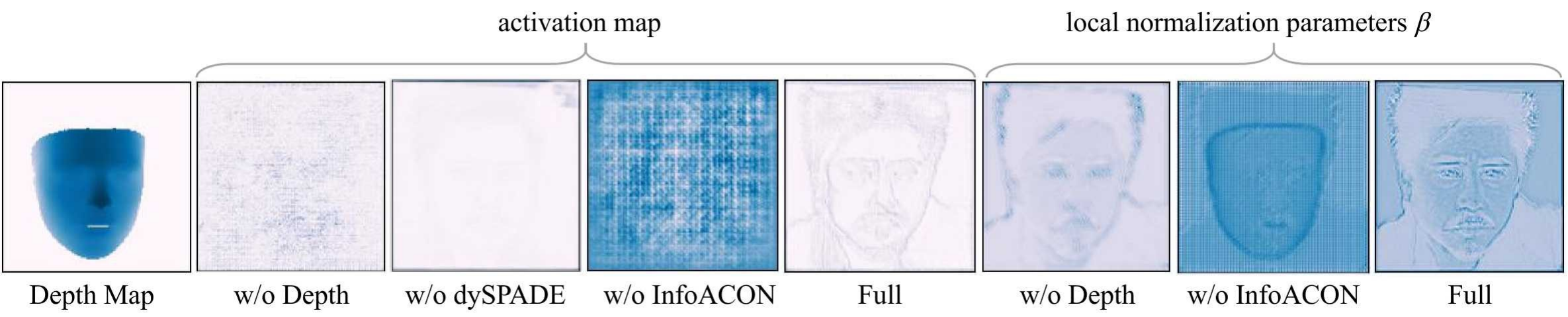}
	\caption{Visualization of activations and adaptation parameters, w.r.t. different model variants of HIDA.}
	\label{fig:vis}
\end{figure*}

\textbf{Quantitative Analysis.}
Table \ref{tab:ablation} lists the performance criteria achieved by these model variants. 
Using depth alone, although improves the geometrical structures (Fig. \ref{fig:exp_ablation}), doesn't consistently contribute to quantitative performance. 
As we gradually add the DySPADE, InfoACON, and DOG modules, both FID and LPIPS consistently decrease. At the same time, Model-C and Model-D achieve comparable SCOOT and FSIM values, in contrast to Model-A. This means that both DySPADE and InfoACON improve the realism of the generated sketch portraits, without significantly changing facial structures. Inspiringly, our full model achieves the best performance in terms of all the quantitative criteria. 

\textbf{Parameter Visualization.} 
We further analyse the impact of each module by removing it from our full model. Fig. \ref{fig:vis} visualizes activation maps and adaptation parameters w.r.t. the corresponding model variants. We can see that depth helps learning effective geometric representations (Full \textit{vs}. w/o Depth). Besides, the proposed dynamic adaptation (DySPADE and InfoACON) boosts the representations, and migrates the artefacts introduced by the incomplete depth map. 
Based on previous analysis, we conclude that our method achieves such inspiring performance, due to a combination of depth and the IDA modules.

\begin{figure}
	\centering
	\includegraphics[width=1\linewidth]{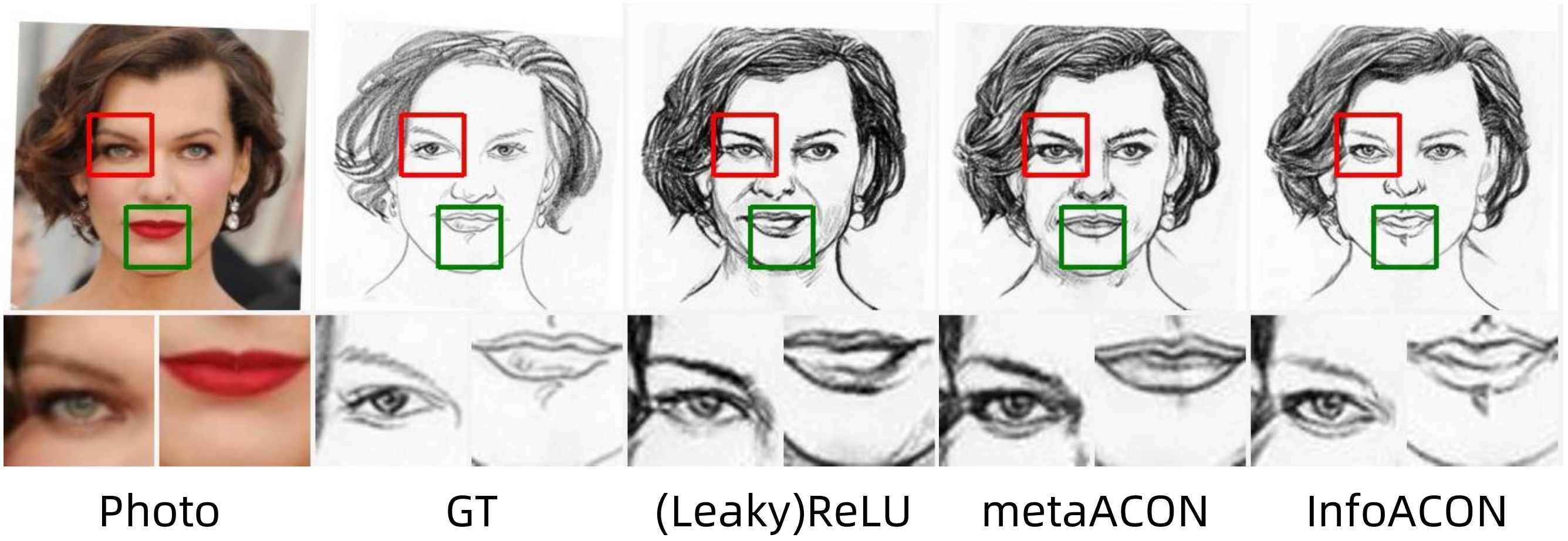}
	\caption{Comparison between activation functions.}
	\label{fig:exp_ACON}
\end{figure}

\textbf{Analysis of InfoACON.}
\label{ssec:exp_ACON}
We further analyze the impacts of dynamic activation functions, including metaACON and the proposed InfoACON. To this end, we build model variants based on Model-B, by (1) using ReLU in the encoder and LeakyReLU in the decoder and discriminator; (2) using metaACON \cite{ma2021acon} in all layers; and (3) using InfoACON in the decoder and metaACON in the other layers.  
As shown in Fig. \ref{fig:exp_ACON}, 
InfoACON makes the generator merely produce distinct sketchy outlines over the mouth region, which is most similar to the ground truth, in terms of style. 
As shown in Table \ref{tab:dinfoACON}, InfoACON achieves the lowest FID and LPIPS, as well as highly comparable SCOOT and FSIM. 
Besides, both metaACON and InfoACON outperform ReLU/LeakyReLU. This means that dynamic activation significantly improves the consistency between the synthesized sketches and those drawn by human artists, in terms of textures. 

\begin{table}
	\centering
	\caption{Comparison between different activation functions.}
	\label{tab:dinfoACON}
		\begin{tabular}{l|cccc}
			\toprule
			&	FID$\downarrow$	&	LPIPS$\downarrow$	&	SCOOT$\uparrow$	&	FSIM$\uparrow$	\\
			\midrule
			(Leaky)ReLU	&	\underline{18.30}	&	0.298	&	0.479	&	\underline{0.539}	\\
			metaACON	&	19.05	&	\underline{0.292}	&	\textbf{0.498}	&	\textbf{0.541}	\\
			InfoACON	&	\textbf{17.41}	&	\textbf{0.291}	&	\underline{0.493}	&	0.536	\\
			\bottomrule
		\end{tabular}
\end{table}

\begin{figure}
	\centering
	\includegraphics[width=\linewidth]{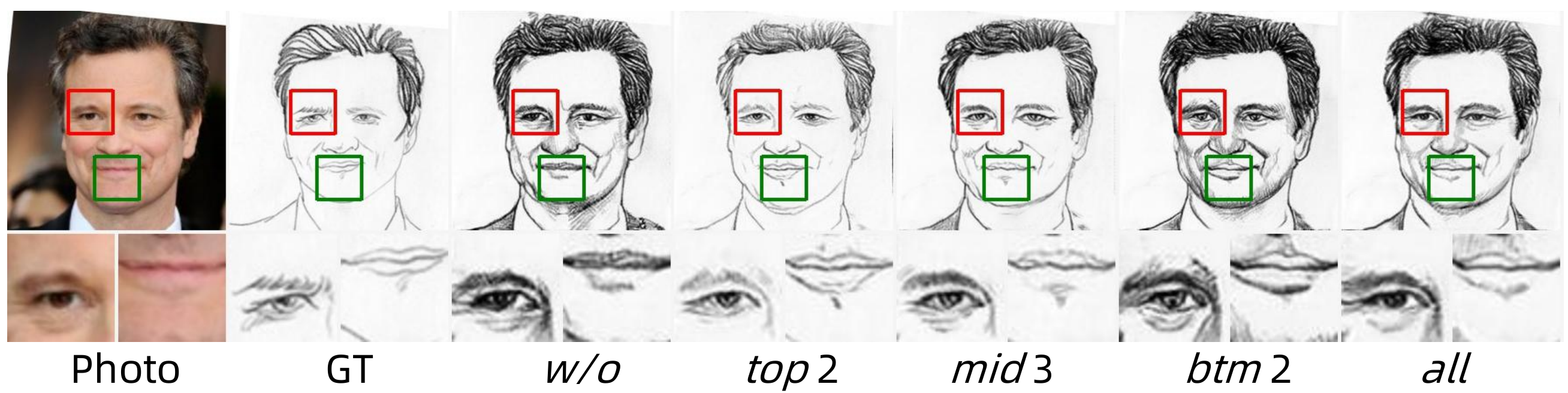}
	\caption{Comparison between different settings of DOG.}
	\label{fig:exp_dfa}
\end{figure}

\textbf{Analysis of DOG.}
\label{ssec:exp_dfa}
In our framework, we apply deformable convolutions only at coarse-scale layers, i.e. the top 2 layers in the decoder. To verify such motivation, we conduct variants of our final model, by using DOG at top 2 layers (\textit{top}2), middle 3 layers (\textit{mid}3), bottom 2 layers (\textit{btm}2), and all layers (\textit{all}), respectively. 
In this experiment, HIDA w/o DOG is the base model. 
%
Fig. \ref{fig:exp_dfa} illustrates the corresponding synthesized sketches. Obviously, the sketches synthesized with DOG present distincter geometric outlines than those without DOG. 
If we apply DOG over the bottom layer, the model fails to generate sketches in Style1. 
This might due to the fact that human painters usually abstract in large areas rather than small ones. 
Besides, the sketch synthesized by \textit{top}2 has the most consistent style compared to the ground truth.
Using DOG over all decoding layers leads to an integrated effects on the synthesized sketch, e.g. confused styles and distinct boundaries. 
We therefore merely use DOG over the top 2 decoding layers in our final model. 

\subsection{Generalization Ability}
\label{ssec:exp_robust}
To evaluate the generalization ability of our framework, we apply the previously learned HIDA model to challenging faces in-the-wild and natural images. Here, we compare with Pix2Pix, SCA-GAN, and GENRE, because the models of MDAL and FSGAN haven't been released. All models are learned from the training set of the FS2K dataset.

\begin{figure}
	\centering
	\includegraphics[width=1\linewidth]{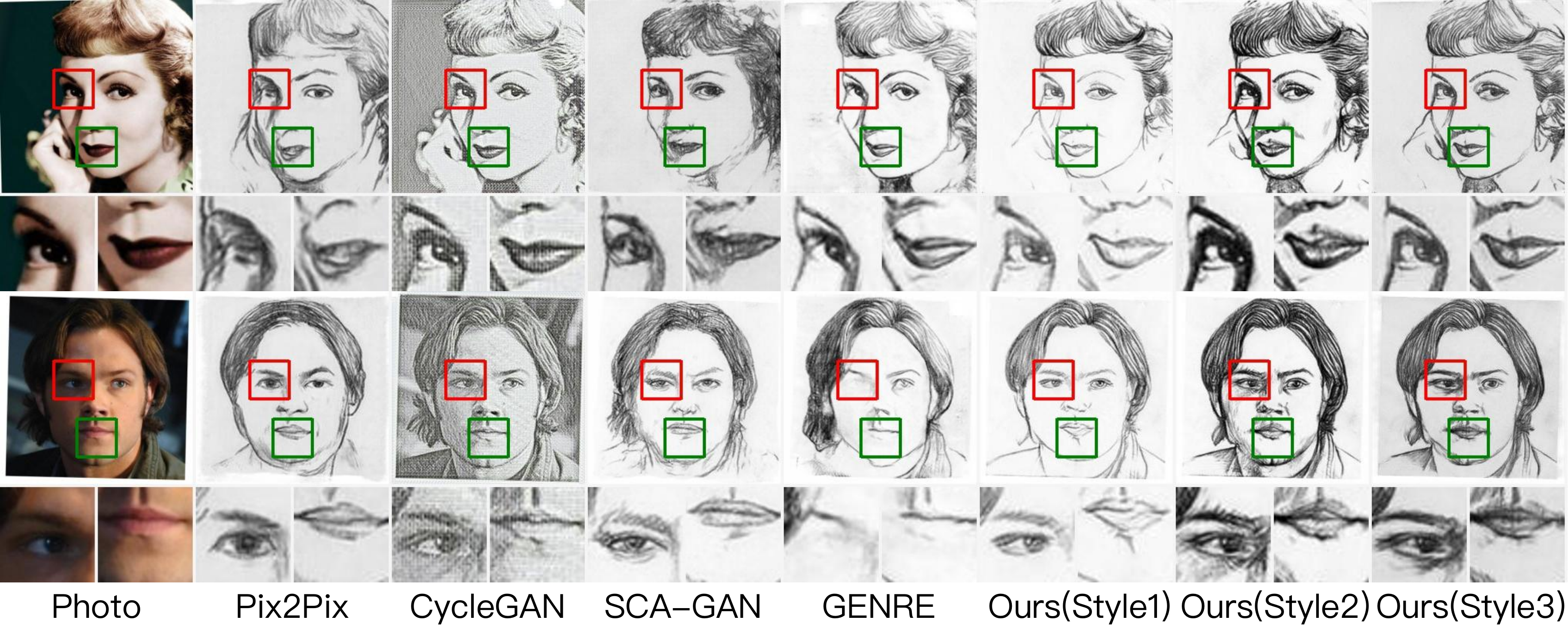}
	\caption{Performance of our method on faces in-the-wild.}
	\label{fig:wild}
\end{figure}

\begin{figure}
	\includegraphics[width=1\linewidth]{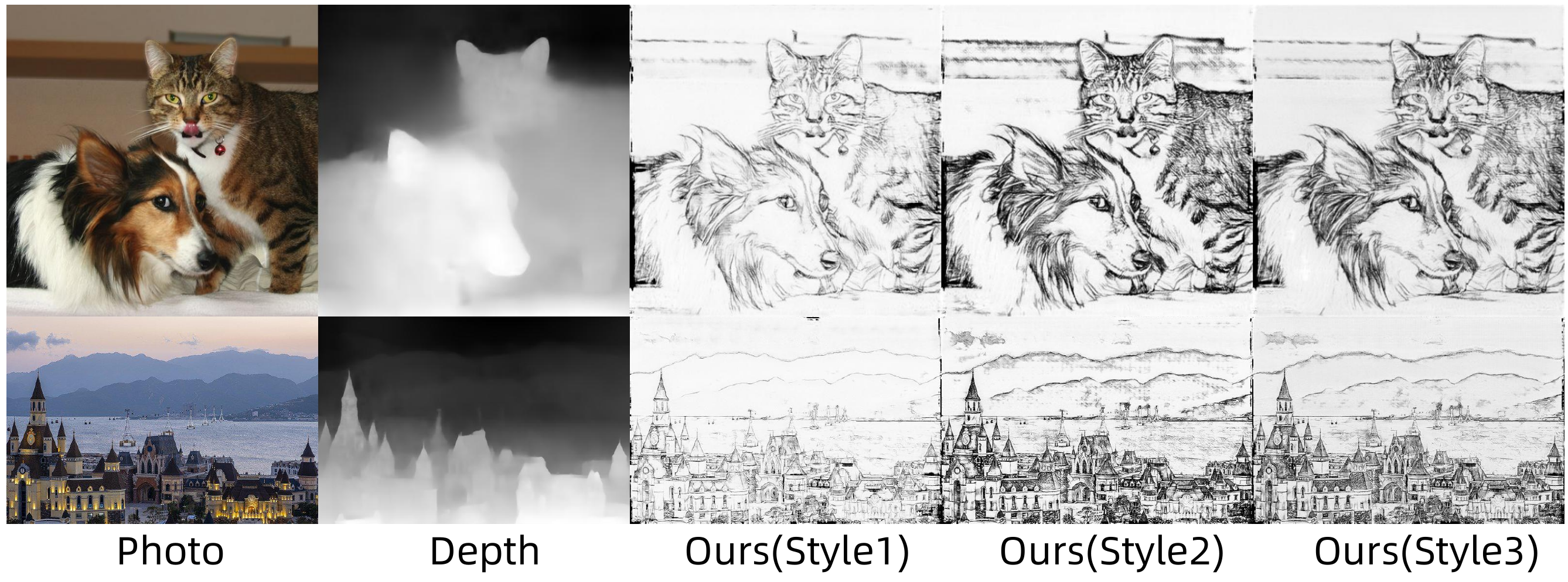}
	\caption{Performance of our model on natural images.}
	\label{fig:natural}
\end{figure}

\textbf{Performance on Faces In-the-wild.}
Fig. \ref{fig:wild} illustrates synthesized sketches on unconstrained faces. These faces have extreme variations in occlusion, pose, lighting, and tone. Generally speaking, our method produces high-quality sketches, in multiple styles, for both examples. Inspiringly, for the example shown in the bottom row, our HIDA model successfully depicts the eyes in shading areas. 
In contrast, the other methods fail to generate some quality details, e.g. eyes of both examples. Moreover, they produce geometric deformations over the mouth of the bottom example. 
Finally, our synthesized sketches vividly characterize the moods shown in the photographic faces.

\textbf{Extension to Natural Images.}
We here apply our previously learned model to several natural images, collected from the Web. Here, we use MiDas \cite{ranftl2020MiDas} instead of 3DDFA for depth estimation. Fig. \ref{fig:natural} shows that our model still produces high-quality sketches, in multiple styles. The synthesized sketches vividly present the geometry and appearance of natural images. 

\begin{figure}
	\centering
	\includegraphics[width=1\linewidth]{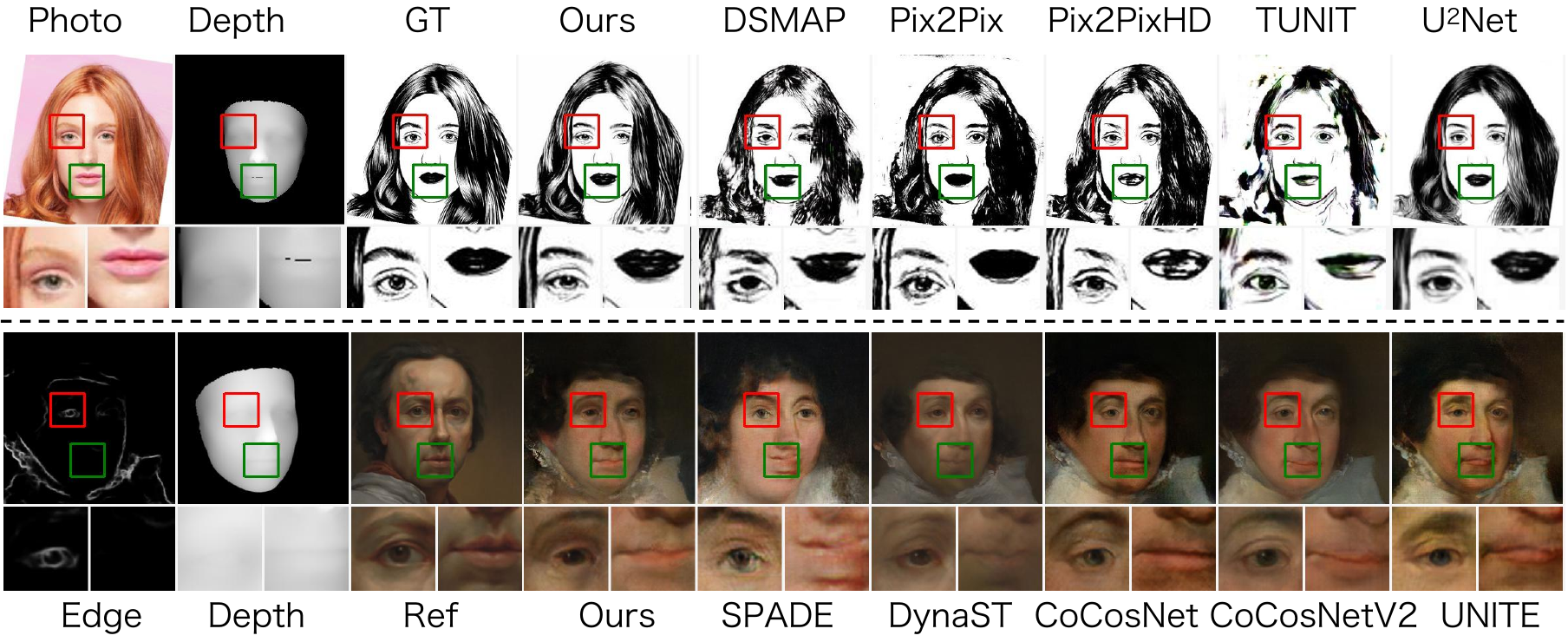}
	\caption{Generated pen-drawings and oil-paintings, on the APDrawing and MetFace datasets.}
	\label{fig:metface}
\end{figure}

\textbf{Extension to other Image-to-Image translation tasks.} 
We additionally apply our method to pen-drawing generation  (with paired data on the APDrawing dataset \cite{YiLLR19}) and exemplar-based image translation (with unpaired data on the MetFace dataset \cite{karras2020training}). 
In the former task, we train and test our full model following standard settings. In the latter task, we use CoCosNet \cite{cocosnet} as the baseline, and modify it by (1) using depth, and (2) replacing the standard SPADE modules in CoCosNet by DySPADE and InfoACON. As shown in Table \ref{tab:addcomp}, our method outperforms previous SOTA methods, in terms of most performance indices. Fig. \ref{fig:metface} shows that our method generates distinct and accurate facial structures, compared to the other methods. Such results demonstrate that the proposed techniques are robust and applicable to other image translation tasks. 


\begin{table}
	\tabcolsep=1.5pt
	\footnotesize
	\centering
	\caption{Comparison with SOTAs on the APDrawing and MetFace datasets.}
	\label{tab:addcomp}
		\begin{tabular}{lcc|lccc}
			\toprule
%
\textbf{APDrawing}	&	FID$\downarrow$		&	LPIPS $\downarrow$	& \textbf{MetaFace} &	FID$\downarrow$	&	LPIPS$\downarrow$ &	Sem.$\uparrow$	\\
			\midrule	
			DSMAP \cite{chang2020domain}	&	71.38	&	0.466 	 & GauGAN \cite{Park2019GauGAN}	&	76.42	&	0.391	 &	0.915	\\
			TUNIT \cite{Baek2021tunit}	&	91.64	&	0.458 &  DyNaST \cite{liu2022dynast}	&	29.25	&	0.375	&	0.917		\\
			Pix2Pix \cite{Isola2017Pix2Pix}	&	80.11	&	0.250  & Cocosnet \cite{cocosnet}	&	34.14	&	0.355	&		0.930		\\
			Pix2PixHD \cite{Wang2017Pix2PixHD}	&	\underline{60.55}	&	\underline{0.206}  & 	CocosnetV2 \cite{zhang2021cocosnetv2}	&	27.98	&	0.296	&	0.939	\\
            U$^2$-Net \cite{qin2020u2net}	&	77.19	&	0.232  & 	UNITE \cite{zhan2021unite}	&	35.91	&	0.356	&	0.930	\\
            HIDA (Ours)	&	\textbf{56.58}	&	\textbf{0.194} 	 &  HIDA (Ours)	&	\textbf{22.62}    &	\textbf{0.174}	&	   \textbf{0.981}  	  \\
			\bottomrule
		\end{tabular}
\end{table}

\section{Conclusions}
\label{sec:conclusion}

In this work, we use comprehensive facial information for synthesizing sketchy portraits. Technically, we propose two informative and dynamic adaptation methods, including a normalization module and an activation function. Extensive experiments show that our method, termed HIDA, can generate high-quality and style-controllable sketches, over a wide range of challenging samples. 
Our work also implies promising applications of dynamic adaptation, or dynamic networks, in more image generation tasks. 
Besides, it is promising to boost the performance of FSS models by combining multi-source datasets. We will explore such works in the near future.

{\small
\bibliographystyle{ieee_fullname}
\bibliography{ref}

\begin{thebibliography}{10}\itemsep=-1pt

\bibitem{Baek2021tunit}
Kyungjune Baek, Yunjey Choi, Youngjung Uh, Jaejun Yoo, and Hyunjung Shim.
\newblock Rethinking the truly unsupervised image-to-image translation.
\newblock In {\em Proceedings of the IEEE/CVF International Conference on
  Computer Vision (ICCV)}, pages 14154--14163, October 2021.

\bibitem{artai22}
Eva Cetinic and James She.
\newblock Understanding and creating art with ai: Review and outlook.
\newblock {\em ACM Trans. Multimedia Comput. Commun. Appl.}, 18(2), feb 2022.

\bibitem{chan2022informativedraw}
Caroline Chan, Fr{\'e}do Durand, and Phillip Isola.
\newblock Learning to generate line drawings that convey geometry and
  semantics.
\newblock In {\em Proceedings of the IEEE/CVF Conference on Computer Vision and
  Pattern Recognition}, pages 7915--7925, 2022.

\bibitem{chang2020domain}
Hsin-Yu Chang, Zhixiang Wang, and Yung-Yu Chuang.
\newblock Domain-specific mappings for generative adversarial style transfer.
\newblock In {\em Computer Vision--ECCV 2020: 16th European Conference,
  Glasgow, UK, August 23--28, 2020, Proceedings, Part VIII 16}, pages 573--589.
  Springer, 2020.

\bibitem{Chen2018Semi}
C Chen, W Liu, X Tan, and KKY Wong.
\newblock Semi-supervised learning for face sketch synthesis in the wild.
\newblock In {\em ACCV}, 2018.

\bibitem{dyrelu}
Yinpeng Chen, Xiyang Dai, Mengchen Liu, Dongdong Chen, Lu Yuan, and Zicheng
  Liu.
\newblock Dynamic relu.
\newblock In {\em European Conference on Computer Vision}, pages 351--367.
  Springer, 2020.

\bibitem{Dai2017Deformable}
Jifeng Dai, Haozhi Qi, Yuwen Xiong, Yi Li, Guodong Zhang, Han Hu, and Yichen
  Wei.
\newblock Deformable convolutional networks.
\newblock In {\em IEEE ICCV}, pages 764--773, 2017.

\bibitem{Fan2021FS2K}
Fan Deng-Ping, Huang Ziling, Zheng Peng, Liu Hong, Qin Xuebin, and Van~Gool
  Luc.
\newblock Deep facial synthesis: A new challenge.
\newblock {\em Machine Intelligence Research}, 19:257–287, 2022.

\bibitem{duan2020multi}
Shuchao Duan, Zhenxue Chen, QM~Jonathan Wu, Lei Cai, and Dan Lu.
\newblock Multi-scale gradients self-attention residual learning for face
  photo-sketch transformation.
\newblock {\em IEEE Transactions on Information Forensics and Security},
  16:1218--1230, 2020.

\bibitem{fan2019scoot}
D. Fan, S. Zhang, Y. Wu, Y. Liu, M. Cheng, B. Ren, Paul~L Rosin, and R. Ji.
\newblock Scoot: A perceptual metric for facial sketches.
\newblock In {\em ICCV}, pages 5612--5622, 2019.

\bibitem{gao2020incremental}
Fei Gao, Jingjie Zhu, Hanliang Jiang, Zhenxing Niu, Weidong Han, and Jun Yu.
\newblock Incremental focal loss gans.
\newblock {\em Information Processing \& Management}, 57(3):102192, 2020.

\bibitem{Gui2020ReviewGAN}
J Gui, Z Sun, Y. Wen, D. Tao, and J. Ye.
\newblock A review on generative adversarial networks: Algorithms, theory, and
  applications.
\newblock {\em IEEE Transactions on Knowledge and Data Engineering}, pages
  1--28, 2021.

\bibitem{3DDFA}
Jianzhu Guo, Xiangyu Zhu, Yang Yang, Fan Yang, Zhen Lei, and Stan~Z Li.
\newblock Towards fast, accurate and stable 3d dense face alignment.
\newblock In {\em Proceedings of the European Conference on Computer Vision
  (ECCV)}, 2020.

\bibitem{han2021dynamic}
Yizeng Han, Gao Huang, Shiji Song, Le Yang, Honghui Wang, and Yulin Wang.
\newblock Dynamic neural networks: A survey.
\newblock {\em IEEE Transactions on Pattern Analysis and Machine Intelligence},
  44(11):7436--7456, 2021.

\bibitem{Heusel2017FID}
M. Heusel, H. Ramsauer, T. Unterthiner, B. Nessler, and S. Hochreiter.
\newblock {GANs} trained by a two time-scale update rule converge to a local
  nash equilibrium.
\newblock In {\em NIPS}, pages 6626--6637. 2017.

\bibitem{FaPN2021ICCV}
Shihua Huang, Zhichao Lu, Ran Cheng, and Cheng He.
\newblock Fapn: Feature-aligned pyramid network for dense image prediction.
\newblock In {\em Proceedings of the IEEE/CVF International Conference on
  Computer Vision (ICCV)}, pages 864--873, October 2021.

\bibitem{Huang2017AdaIN}
X. Huang and S. Belongie.
\newblock Arbitrary style transfer in real-time with adaptive instance
  normalization.
\newblock In {\em ICCV}, pages 1510--1519, oct 2017.

\bibitem{Isola2017Pix2Pix}
P. Isola, J.~Y. Zhu, T. Zhou, and A.~A. Efros.
\newblock Image-to-image translation with conditional adversarial networks.
\newblock In {\em CVPR}, pages 1125--1134, 2017.

\bibitem{karras2020training}
Tero Karras, Miika Aittala, Janne Hellsten, Samuli Laine, Jaakko Lehtinen, and
  Timo Aila.
\newblock Training generative adversarial networks with limited data.
\newblock {\em Advances in neural information processing systems},
  33:12104--12114, 2020.

\bibitem{Karras2018StyleGAN}
T. Karras, S. Laine, and T. Aila.
\newblock A style-based generator architecture for generative adversarial
  networks.
\newblock In {\em CVPR}, June 2019.

\bibitem{li2021genre}
Xiang Li, Fei Gao, and Fei Huang.
\newblock High-quality face sketch synthesis via geometric normalization and
  regularization.
\newblock In {\em 2021 IEEE International Conference on Multimedia and Expo
  (ICME)}, pages 1--6. IEEE, 2021.

\bibitem{li2019im2pencil}
Yijun Li, Chen Fang, Aaron Hertzmann, Eli Shechtman, and Ming-Hsuan Yang.
\newblock Im2pencil: Controllable pencil illustration from photographs.
\newblock In {\em Proceedings of the IEEE/CVF Conference on Computer Vision and
  Pattern Recognition}, pages 1525--1534, 2019.

\bibitem{liu2022dynast}
Songhua Liu, Jingwen Ye, Sucheng Ren, and Xinchao Wang.
\newblock Dynast: Dynamic sparse transformer for exemplar-guided image
  generation.
\newblock In {\em European Conference on Computer Vision}, pages 72--90.
  Springer, 2022.

\bibitem{lv2021learning}
Zhengyao Lv, Xiaoming Li, Xin Li, Fu Li, Tianwei Lin, Dongliang He, and
  Wangmeng Zuo.
\newblock Learning semantic person image generation by region-adaptive
  normalization.
\newblock In {\em Proceedings of the IEEE/CVF Conference on Computer Vision and
  Pattern Recognition}, pages 10806--10815, 2021.

\bibitem{ma2021acon}
Ningning Ma, Xiangyu Zhang, Ming Liu, and Jian Sun.
\newblock Activate or not: Learning customized activation.
\newblock In {\em Proceedings of the IEEE/CVF Conference on Computer Vision and
  Pattern Recognition}, pages 8032--8042, 2021.

\bibitem{nie2021unconstrained}
Lin Nie, Lingbo Liu, Zhengtao Wu, and Wenxiong Kang.
\newblock Unconstrained face sketch synthesis via perception-adaptive network
  and a new benchmark.
\newblock {\em arXiv preprint arXiv:2112.01019}, 2021.

\bibitem{Park2019GauGAN}
T. Park, M.-Y. Liu, T.-C. Wang, and J.-Y. Zhu.
\newblock Semantic image synthesis with spatially-adaptive normalization.
\newblock In {\em CVPR}, pages 2337--2346, June 2019.

\bibitem{Peng2019DeepPGM}
C. {Peng}, N. {Wang}, J. {Li}, and X. {Gao}.
\newblock Face sketch synthesis in the wild via deep patch representation-based
  probabilistic graphical model.
\newblock {\em IEEE TIFS}, 15:172--183, 2020.

\bibitem{qi2022biphasic}
Xingqun Qi, Muyi Sun, Qi Li, and Caifeng Shan.
\newblock Biphasic face photo-sketch synthesis via semantic-driven generative
  adversarial network with graph representation learning.
\newblock {\em arXiv preprint arXiv:2201.01592}, 2022.

\bibitem{qin2020u2net}
Xuebin Qin, Zichen Zhang, Chenyang Huang, Masood Dehghan, Osmar~R Zaiane, and
  Martin Jagersand.
\newblock U2-net: Going deeper with nested u-structure for salient object
  detection.
\newblock {\em Pattern Recognition}, 106:107404, 2020.

\bibitem{ranftl2020MiDas}
Ren{\'e} Ranftl, Katrin Lasinger, David Hafner, Konrad Schindler, and Vladlen
  Koltun.
\newblock Towards robust monocular depth estimation: Mixing datasets for
  zero-shot cross-dataset transfer.
\newblock {\em IEEE transactions on pattern analysis and machine intelligence},
  2020.

\bibitem{tan2021diverse}
Zhentao Tan, Menglei Chai, Dongdong Chen, Jing Liao, Qi Chu, Bin Liu, Gang Hua,
  and Nenghai Yu.
\newblock Diverse semantic image synthesis via probability distribution
  modeling.
\newblock In {\em Proceedings of the IEEE/CVF Conference on Computer Vision and
  Pattern Recognition}, pages 7962--7971, 2021.

\bibitem{ulyanov2017IN}
D. Ulyanov, A. Vedaldi, and V. Lempitsky.
\newblock Improved texture networks: Maximizing quality and diversity in
  feed-forward stylization and texture synthesis.
\newblock pages 4105--4113, 2017.

\bibitem{Wang2013Transductive}
N. Wang, D. Tao, X. Gao, X. Li, and J. Li.
\newblock Transductive face sketch-photo synthesis.
\newblock {\em IEEE TNNLS}, 24(9):1364--1376, 2013.

\bibitem{Wang2017Pix2PixHD}
T.~C. Wang, M.~Y. Liu, J.~Y. Zhu, A. Tao, J. Kautz, and B. Catanzaro.
\newblock High-resolution image synthesis and semantic manipulation with
  conditional {GANs}.
\newblock pages 8798--8807, 2018.

\bibitem{YiLLR19}
Ran Yi, Yong-Jin Liu, Yu-Kun Lai, and Paul~L Rosin.
\newblock {APDrawingGAN}: Generating artistic portrait drawings from face
  photos with hierarchical gans.
\newblock In {\em {IEEE} Conference on Computer Vision and Pattern Recognition
  (CVPR '19)}, pages 10743--10752, 2019.

\bibitem{gao2020cagan}
J. Yu, S. Shi, F. Gao, D. Tao, and Q. Huang.
\newblock Towards realistic face photo-sketch synthesis via composition-aided
  {GANs}.
\newblock {\em IEEE TCYB}, 51(9):4350--4362, 2021.

\bibitem{zhan2021unite}
Fangneng Zhan, Yingchen Yu, Kaiwen Cui, Gongjie Zhang, Shijian Lu, Jianxiong
  Pan, Changgong Zhang, Feiying Ma, Xuansong Xie, and Chunyan Miao.
\newblock Unbalanced feature transport for exemplar-based image translation.
\newblock In {\em Proceedings of the IEEE/CVF Conference on Computer Vision and
  Pattern Recognition}, pages 15028--15038, 2021.

\bibitem{Zhang2011FSIM}
L. Zhang, L. Zhang, X. Mou, and D. Zhang.
\newblock {FSIM}: a feature similarity index for image quality assessment.
\newblock {\em IEEE TIP}, 20(8):2378--2386, 2011.

\bibitem{Zhang2019TIP}
M. {Zhang}, R. {Wang}, X. {Gao}, J. {Li}, and D. {Tao}.
\newblock Dual-transfer face sketch–photo synthesis.
\newblock {\em TIP}, 28(2):642--657, Feb 2019.

\bibitem{cocosnet}
Pan Zhang, Bo Zhang, Dong Chen, Lu Yuan, and Fang Wen.
\newblock Cross-domain correspondence learning for exemplar-based image
  translation.
\newblock In {\em Proceedings of the IEEE/CVF Conference on Computer Vision and
  Pattern Recognition}, pages 5143--5153, 2020.

\bibitem{zhang2018lpips}
R. Zhang, P. Isola, A.~A Efros, E. Shechtman, and O. Wang.
\newblock The unreasonable effectiveness of deep features as a perceptual
  metric.
\newblock In {\em CVPR}, pages 586--595, 2018.

\bibitem{Zhang2018IJCAI}
S. Zhang, R. Ji, J. Hu, Y. Gao, and Lin C.-W.
\newblock Robust face sketch synthesis via generative adversarial fusion of
  priors and parametric sigmoid.
\newblock In {\em IJCAI}, pages 1163--1169, 2018.

\bibitem{Zhang2019MAL}
S. {Zhang}, R. {Ji}, J. {Hu}, X. {Lu}, and X. {Li}.
\newblock Face sketch synthesis by multidomain adversarial learning.
\newblock {\em IEEE TNNLS}, 30(5):1419--1428, May 2019.

\bibitem{zhang2021cocosnetv2}
Xingran Zhou, Bo Zhang, Ting Zhang, Pan Zhang, Jianmin Bao, Dong Chen, Zhongfei
  Zhang, and Fang Wen.
\newblock Cocosnet v2: Full-resolution correspondence learning for image
  translation.
\newblock In {\em Proceedings of the IEEE/CVF Conference on Computer Vision and
  Pattern Recognition (CVPR)}, pages 11465--11475, 2021.

\bibitem{Zhu2017CycleGAN}
J. Zhu, T. Park, P. Isola, and A.~A Efros.
\newblock Unpaired image-to-image translation using cycle-consistent
  adversarial networks.
\newblock In {\em ICCV}, pages 2242--2251, 2017.

\bibitem{zhu2021sketch}
Mingrui Zhu, Changcheng Liang, Nannan Wang, Xiaoyu Wang, Zhifeng Li, and Xinbo
  Gao.
\newblock A sketch-transformer network for face photo-sketch synthesis.
\newblock In {\em International Joint Conference on Artificial Intelligence},
  2021.

\bibitem{zhu2020sean}
Peihao Zhu, Rameen Abdal, Yipeng Qin, and Peter Wonka.
\newblock Sean: Image synthesis with semantic region-adaptive normalization.
\newblock In {\em Proceedings of the IEEE/CVF Conference on Computer Vision and
  Pattern Recognition}, pages 5104--5113, 2020.

\end{thebibliography}
}

\end{document}